\documentclass[lettersize,journal]{IEEEtran}
\usepackage{amsmath,amsfonts}
\usepackage{algorithmic}
\usepackage{algorithm}
\usepackage{array}
\usepackage[caption=false,font=normalsize,labelfont=sf,textfont=sf]{subfig}
\usepackage{textcomp}
\usepackage{stfloats}
\usepackage{url}
\usepackage{verbatim}
\usepackage{graphicx}
\usepackage{cite}

\usepackage{booktabs}
\usepackage{colortbl}
\usepackage{color}
\usepackage{amsfonts}
\usepackage[table,dvipsnames]{xcolor}
\usepackage{multirow}

\newcommand{\blue}[1]{\textcolor{black}{#1}}
\newcommand{\red}[1]{\textcolor{red}{#1}}
\newcommand{\green}[1]{\textcolor{green}{#1}}
\newcommand{\pink}[1]{\textcolor[RGB]{220,27,116}{#1}}
\newcommand{\anno}[1]{\textcolor[RGB]{65,128,128}{#1}}

\begin{document}

\title{Spot-adaptive Knowledge Distillation}

\author{Jie Song,~\IEEEmembership{{Member},~IEEE},~Ying Chen,~Jingwen Ye,~Mingli Song,~\IEEEmembership{{Senior~Member},~IEEE}% <-this % stops a space
\thanks{This work is funded by the National Key R\&D Program of China (Grant No: 2018AAA0101503) and the Science and technology project of SGCC (State Grid Corporation of China): fundamental theory of human-in-the-loop hybrid-augmented intelligence for power grid dispatch and control.}
\thanks{J. Song, Y. Chen, J. Ye and M. Song are with the Zhejiang University, Hangzhou, China. Email: sjie@zju.edu.cn; lynesychen@zju.edu.cn; yejingwen@zju.edu.cn; brooksong@zju.edu.cn}
%\thanks{Manuscript received April 19, 2021; revised August 16, 2021.}
}

% The paper headers
\markboth{Journal of \LaTeX\ Class Files,~Vol.~14, No.~8, August~2021}%
{Shell \MakeLowercase{\textit{et al.}}: A Sample Article Using IEEEtran.cls for IEEE Journals}

% \IEEEpubid{0000--0000/00\$00.00~\copyright~2021 IEEE}
% Remember, if you use this you must call \IEEEpubidadjcol in the second
% column for its text to clear the IEEEpubid mark.

\maketitle

\begin{abstract}
Knowledge distillation~(KD) has become a well established paradigm for compressing deep neural networks. The typical way of conducting knowledge distillation is to train the student network under the supervision of the teacher network to harness the knowledge at one or multiple spots~(\textit{i.e.}, layers) in the teacher network. The distillation spots, once specified, will not change for all the training
samples, throughout the whole distillation process. In this work, we argue that distillation spots should be adaptive to training samples and distillation epochs. We thus propose a new distillation strategy, termed spot-adaptive KD~(SAKD), to adaptively determine the distillation spots in the teacher network per sample, at every training iteration during the whole distillation period. As SAKD actually focuses on ``where to distill'' instead of ``what to distill'' that is widely investigated by most existing works, it can be seamlessly integrated into existing distillation methods to further improve their performance. Extensive experiments with 10 state-of-the-art distillers are conducted to demonstrate the effectiveness of SAKD for improving their distillation performance, under both homogeneous and heterogeneous distillation settings. Code is available at \url{https://github.com/zju-vipa/spot-adaptive-pytorch}.
\end{abstract}


\section{Introduction}

With the rapid development of deep learning in the last decade, deep neural networks~(DNNs) have become the predominant models in various fields, including computer vision, natural language processing~(NLP), \textit{etc}. However, for a long time till today, DNNs have always been criticized as being excessively large to be deployed to resource-limited edge devices. To make DNNs more applicable in these real-world scenarios, knowledge distillation~(KD)~\cite{Hinton2015DistillingTK,tian2019crd} has been proposed for crafting the lightweight substitutes for these expensive DNNs. The main idea is adopting a teacher-student learning scheme, where the competitive lightweight substitutes, called \textit{students}, are produced by mimicking some behaviors from well-behaved yet cumbersome DNNs which play the role of \textit{teachers}. By harnessing the dark knowledge learned by teacher models, the lightweight student models are expected to achieve comparable performance, yet with much fewer parameters.
\begin{figure}[t]
    \centering
    \includegraphics[scale=0.43]{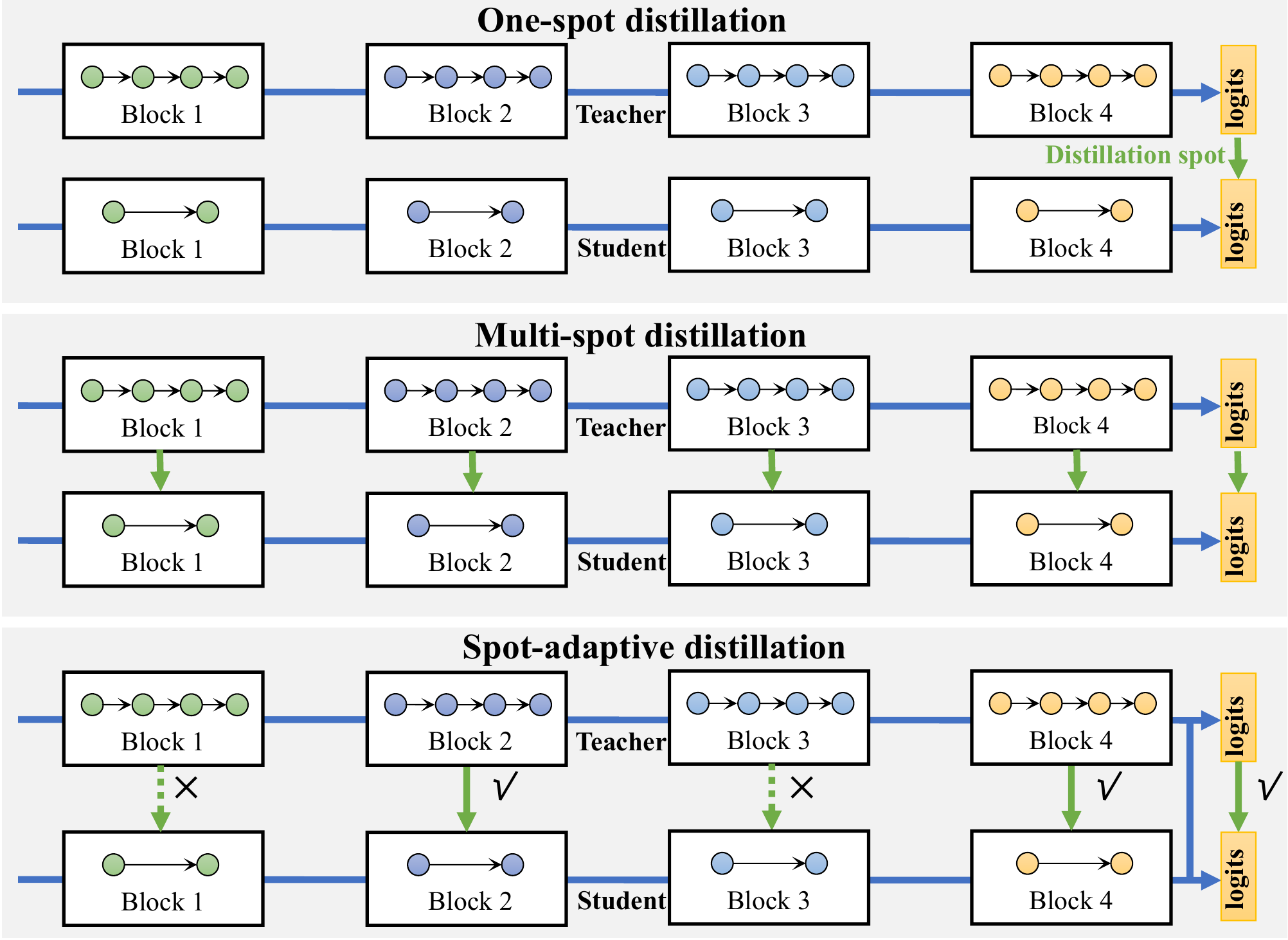}
    \caption{A schematic illustration of one-spot distillation, multi-spot distillation and the proposed spot-adaptive distillation. The dotted green arrows denote that these distillation spots are adaptively excluded dependent on samples and training iterations.}
    \label{fig:one-vs-multi}
\end{figure}
\vspace{1.0pt}

After recent years of development, KD has made remarkable progress and nowadays become a well established paradigm for compressing DNNs. Based on the number of distillation spots, existing distillation methods can be roughly categorized into two schools: \textit{one-spot} distillation~\cite{Hinton2015DistillingTK,tian2019crd,Park2019RelationalKD,Cho2019OnTE,Yuan_2020_CVPR}, and \textit{multi-spot} distillation~\cite{Romero2015FitNetsHF,Zagoruyko2017AT,Yim2017AGF,Kim2018ParaphrasingCN,Heo2019KnowledgeTV,pmlr-v80-srinivas18a}, as shown in Fig.~\ref{fig:one-vs-multi}. In one-spot distillation, KD happens at only one layer, typically the logit layer, in the teacher model. For example, Hinton \textit{et al.}~\cite{Hinton2015DistillingTK} proposed to minimize the KL divergence between the probabilistic outputs of the teacher and the student networks.  Contrastive representation distillation~(CRD)~\cite{tian2019crd} adopts a contrastive learning approach to distill structural knowledge, \textit{i.e.}, inter-dependencies between different output dimensions of the representation layer. Relational knowledge distillation~(RKD)~\cite{Park2019RelationalKD} transfers mutual relations of data examples from teacher models to student models, where the mutual relations are produced within a single representation layer. As DNNs are typically composed of dozens of layers, researchers have also investigated how to leverage more layers to boost the performance of KD. Multi-spot distillation provides student models with more supervision signals by mining knowledge from multiple layers in the teacher network. For example, FitNets~\cite{Romero2015FitNetsHF} adopts not only the outputs but also the intermediate representations learned by the teacher as hints for training the student. Attention transfer~\cite{Zagoruyko2017AT} improves the performance of a student CNN network by forcing it to mimic the attention maps at different layers of a powerful teacher network. Activation boundary~(AB)~\cite{Heo2019KnowledgeTV} transfers the knowledge in teacher models via distilling activation boundaries formed by hidden neurons at different layers. As multi-spot distillation methods utilize more information from teacher models than one-spot counterparts, they are usually deemed to exhibit superior performance for training student models.

%once the distillation spots are determined, they remains fixed for all the training samples, throughout the whole distillation process. However, as different training samples are not equally valuable for distillation, the optimal distillation spots per sample for the training data may be somewhat different accordingly. Moreover, as the same sample plays different roles in different stages of the distillation process, the optimal distillation spots per sample should not be fixed during the distillation process. Existing distillation methods overlooked these issues and simply adopt a never-changing distillation strategy for all the samples throughout the whole distillation process, which hinders their further performance improvement.
Existing distillation methods, whether they are one-spot or multi-spot, share one common characteristic:
the distillation spot is usually a manual design choice which \blue{is} inefficient to optimize for, especially for networks with hundreds or thousands of layers. On the one hand, if the distillation spots are set too sparsely, the student model \blue{is} not sufficiently supervised by the teacher. On the other hand, if the distillation spots are set too densely, \textit{e.g.}, every possible layer or neuron, the student model learning can be over-regularized, which also deteriorates distillation performance. Furthermore, current methods employ a \textit{global distillation} strategy, \textit{i.e.}, the distillation spots are fixed for all the samples once specified. The underlying assumption is that such distillation spots are optimal for the entire data distribution, which is not true in many cases.
%For example, for \blue{some hard samples that the teacher makes wrong predictions}, the wrong predictions are harmful for distillation while the intermediate features can be helpful. For some other samples, the intermediate features can be noisy for distillation while the teacher still makes correct final predictions due to the layer-by-layer abstraction which gradually extracts useful features by discarding noisy features.
Ideally, we would like the distillation spots to be determined automatically for each possible spot, per sample.

In this work, we propose a new distillation strategy, termed spot-adaptive KD~(SAKD), to make the distillation spots adaptive to training samples and distillation stages. To this end, we first merge the student model and the teacher model into a multi-path routing network, as shown in Fig.~\ref{fig:overview}. When the data flow through the network, there are many feasible paths for the data to reach the output layer. A lightweight policy network is devised to make decisions per sample on the optimal propagation path when data reach the branch spots in the network. If the data are routed to layers of the teacher model by the policy network, it indicates that the counterpart layers in the student model~(abbreviated to \textit{student layers}) can not yet replace the layers in the teacher model~(abbreviated to \textit{teacher layers}). The knowledge in these teacher layers thus should be distilled into corresponding student layers. Otherwise, if the data are routed to some student layers by the policy network, it indicates that these student layers are good substitutes for corresponding teacher layers, yielding superior or at least comparable performance. In this case, the distillation is not allowed in these layers. As the policy network is devised on top of the routing network and is optimized simultaneously with the routing network, it can automatically determine the optimal distillation spots per sample, at different training iterations of the student model.

The proposed method focuses on ``where to distill'', which is vastly different from and orthogonal to current literature where ``what to distill'' is mainly investigated. It thus can be seamlessly combined with existing methods to further enhance their distillation performance. Specifically, the proposed method is naturally compatible with homogeneous distillation where the student model is in the same-style architecture as the teacher model. However, experiments demonstrate that the proposed method also works surprisingly well under heterogeneous distillation settings where the student model differs largely from that of the teacher model. Moreover, although the proposed method is designed primarily for multi-spot distillation, it can also boost the performance of one-spot distillation by dynamically determining distillation or not for each training sample.

In a nutshell, we made following three main contributions in this work:
\begin{itemize}
    \item To our best knowledge, we are the first to introduce the adaptive distillation problem where the distillation spots should be adaptive to different training samples and varying distillation stages.
    \item We propose a novel spot-adaptive distillation strategy to automatically determine the distillation spots, making the distillation spots adaptive to the training samples and distillation stages.
    \item Extensive experiments under various experimental settings are conducted to showcase the effectiveness of the proposed method for improving distillation performance of existing state-of-the-art distillers.
\end{itemize}

The reminder of this work is organized as follows. Section~\ref{sec:related-work} presents an overview of related works on knowledge distillation and routing networks. Section~\ref{sec:SAKD} describes the proposed spot-adaptive distillation in detail. The experimental setups and results are provided in Section~\ref{sec:experiments}.The conclusions are drawn in Section~\ref{sec:conclusion}.

\section{Related Work}
\label{sec:related-work}
%We briefly review two research topics which are most related to this work, including knowledge distillation and routing network that also involves selecting the optimal route in a multi-path network.
\subsection{Knowledge Distillation}
Knowledge distillation has attracted widespread attention due to its importance in deploying DNNs to low-capacity edge devices. The main idea is leveraging the \textit{dark knowledge} in a bulky teacher to craft a compact student model, of which the performance is expected to be on par with the teacher. Over the last several years, most works devote themselves to the exploration of ``what to distill'', \textit{i.e.}, forms of knowledge for distillation. Representative forms include soft targets~\cite{Hinton2015DistillingTK}, features~\cite{Romero2015FitNetsHF,song2021tree}, attention~\cite{Zagoruyko2017AT}, factors~\cite{Kim2018ParaphrasingCN}, activation boundary~\cite{Heo2019KnowledgeTV}, sample relationship~\cite{Liu2019KnowledgeDV,Park2019RelationalKD,Tung2019SimilarityPreservingKD} and so on. By imitating the teacher to behave in a similar way, the student model achieves comparable performance to the teacher model even with much fewer parameters. \blue{In this work, we focus on ``where to distill'' instead of ``what to distill'', which is largely overlooked by existing works. Moreover, as ``where to distill'' is orthogonal to ``what to distill'',  the proposed method can be seamlessly integrated into existing distillation methods to further improve their performance.}

\begin{figure*}[!ht]
    \centering
    \includegraphics[scale=0.53]{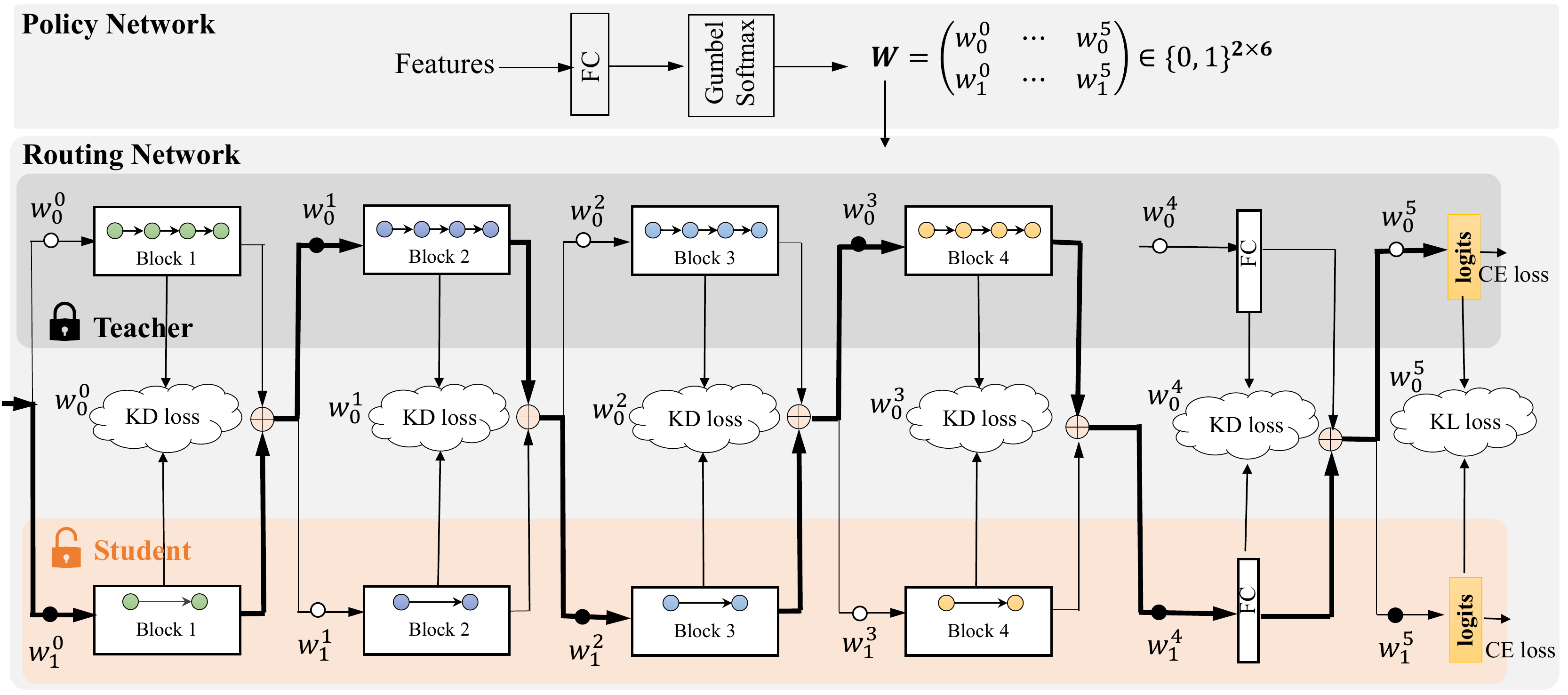}
    \caption{An illustrative diagram of the proposed method. The upper is the policy network and the bottom is the routing network. In this example, there are six possible distillation spots. The input of the policy network is the features from the teacher and the student models. The output of the policy network determines the spots for distillation. The teacher is fixed during distillation.}
    \label{fig:overview}
\end{figure*}

Existing KD methods can be divided into different categories according to different taxonomies. Based on the number of spots where distillation happens, we roughly divide existing KD methods into one-spot~\cite{Hinton2015DistillingTK,tian2019crd,Park2019RelationalKD,Cho2019OnTE,Yuan_2020_CVPR} and multi-spot~\cite{Romero2015FitNetsHF,Zagoruyko2017AT,Yim2017AGF,Kim2018ParaphrasingCN,Heo2019KnowledgeTV,pmlr-v80-srinivas18a} distillation. \blue{The proposed distiller is a spot-adaptive approach where the number of spots for distillation is automatically determined rather than pre-defined}. Based on the similarity between the teacher and the student in architectures, KD can also be categorized into homogeneous distillation~\cite{Yim2017AGF,tian2019crd} and heterogeneous distillation~\cite{Jin2019KnowledgeDV,tian2019crd,touvron2021training}, where the heterogeneous distillation is deemed more challenging due to the large architecture difference~\cite{tian2019crd}. \blue{The proposed method in this work makes few assumptions about the architectures of the teacher and the student, and thus it can be applied to both homogeneous and heterogeneous distillation scenarios}. KD has derived a variety of different problem settings except the canonical teacher-student distillation, including mutual distillation~\cite{Zhang2018DeepML,Yao2020DCM}, self-distillation~\cite{Zhang2019BeYO,Yuan2020RevisitingKD,NEURIPS2020_1731592a}, knowledge amalgamation~\cite{Ye2019StudentBT,Shen2019AmalgamatingKT} and data-free distillation~\cite{chen2019data,fang2019data,fu2021elastic,fang2021mosaicking}. In mutual distillation, an ensemble of student models learn collaboratively and teach each other throughout the training process, enabling the teacher and the student models progress together. Self distillation assumes no teacher network available and distills knowledge within the student network itself by using outputs of deeper layers to supervise shallower layers. Knowledge amalgamation aims to craft a single multi-task student model by fusing the knowledge from multiple teacher models. Data-free distillation relaxes the assumption that the training data of the teacher model is available for training the student, \textit{i.e.}, the knowledge should be distilled to the student without any original training data. \blue{In this work, we follow the canonical distillation settings where both the teacher network and the original training data are available for training the student network. However, we believe the general idea can be also applied to various KD settings, which is left for future work.}

\subsection{Routing Networks}
Routing networks~\cite{Rosenbaum2018RoutingNA,Rosenbaum2019RoutingNA,NEURIPS2018_310ce61c} are a type of neural networks with high modularity, which is a key property required for encouraging task decomposition, reducing model complexity, and improving model generalization. A routing network usually consists of two trainable components: a set of function modules and a policy agent. In neural network settings, the function modules are implemented by sub-networks and used as candidate modules for processing the input data. For each sample, the policy agent selects a subset of function modules from these candidates, assembles them into a complete model, and applies the assembled model to the input data for task predictions. Several algorithms have been proposed for optimizing the policy module, including genetic algorithm~\cite{Fernando2017PathNetEC}, multi-agent reinforcement learning~\cite{Rosenbaum2018RoutingNA}, reparameterization strategies~\cite{Rosenbaum2019RoutingNA}, \textit{etc}.

Routing networks are closely related to conditional computation~\cite{Bengio2015ConditionalCI,Stafford2018ACCMEA}, mixture of experts~\cite{Jacobs1991AdaptiveMO}, as well as their modern attention based~\cite{Riemer2016CorrectingFW} and sparse~\cite{Shazeer2017OutrageouslyLN} variants. They have been successfully applied to several fields like multi-task learning~\cite{Rosenbaum2018RoutingNA}, transfer learning~\cite{Guo2019SpotTuneTL} and language modeling~\cite{NEURIPS2018_310ce61c}. In this work, with the aid of a routing network, we propose a novel distillation strategy to automatically determine the spots of distillation in the network.

\section{Spot-adaptive Knowledge Distillation}
\label{sec:SAKD}
In this section, we introduce the proposed spot-adaptive KD. The overview of the proposed method is shown in Fig.~\ref{fig:overview}. The whole model consists of two main components: a multi-path routing network and a lightweight policy network. The multi-path routing network is composed by the teacher model and the student model, with the adaption layers to adapt their features to each other if necessary. The policy network is used to make routing decisions per sample on the data flow path when the data reach the branch spots in the routing network.

The general idea of the proposed distillation method is to automatically determine whether or not to conduct the distillation at the candidate distillation spots, as shown in Fig.~\ref{fig:overview}.  If the samples are routed to certain teacher layers by the policy network, it indicates that the counterpart student layers can not yet replace these teacher layers. The knowledge in these teacher layers thus should be distilled into corresponding student layers. Otherwise, if the data are transmitted to some student layers by the policy network, it indicates that these student layers are good substitutes for corresponding teacher layers, yielding superior or at least comparable performance. No distillation is needed any more in this case. The ultimate goal of distillation is to make the policy network gradually choose the student layers for routing the data, which implies the student model is a good substitute for the teacher network.

\subsection{The Multi-path Routing Network}
Without loss of generality, assume a canonical Convolution Neural Network~(CNN) for visual classification is composed of several convolution blocks for representation learning, a fully connected layer for vectorizing the feature maps, and a softmax layer for making probability predictions. Each convolution block consists of several convolution layers, each followed by a non-linear activation layer and a batch normalization~\cite{pmlr-v37-ioffe15} layer\footnote{Different architectures may have different network configurations.}. Generally speaking, after each block, the feature maps are downsized by factor of 2 or more with a pooling layer or a convolution layer.
Formally, we denote the function underlying a teacher CNN by
\begin{equation}
    \mathcal{T}=S\circ F^t \circ B^t_{N}\circ\cdot\cdot\cdot\circ B^t_1,
\end{equation}
and the function underlying a student CNN by
\begin{equation}
    \mathcal{S}=S\circ F^s \circ B^s_{N}\circ\cdot\cdot\cdot\circ B^s_1,
\end{equation}
where $S$ denotes the softmax function, and $F$ the linear function. $B_i$ denotes the function underlying the $i$-th block. The superscript $s$ and $t$ denote the student and the teacher models, respectively. The symbol $\circ$ denotes the function composition operation. The multi-path routing network is composed by the teacher and the student networks, with their intermediate layers interconnected with each other. However, there may exist dimension mismatch between the teacher and the student features. We adopt adaption layers~\cite{Romero2015FitNetsHF}, which are implemented by $1\times1$ convolution layers, to align their features. The underlying function of the multi-path routing network is denoted by
\begin{equation}
\label{eq:super-net}
    \mathcal{M}= S\circ \hat{F}\circ\hat{B}_N\circ\cdot\cdot\cdot\circ\hat{B}_1,\ \ \ \ \
\end{equation}
\begin{equation}
\label{eq:fusing}
    \left\{
    \begin{aligned}
        &\hat{F}=wF^t+(1-w)F^s,\\
    &\hat{B}_i=w_iB^t_i+(1-w_i)B^s_i,   \ 1\le i \le N,
    \end{aligned}
    \right.
\end{equation}
where $w$ and $w_i$ are the feature fusing weights produced by the policy network, bounded by $\left[0, 1\right]$. When the feature fusing weights take discrete values from $\{0, 1\}$, the network actually becomes a combinatorial network whose layers are composed of intertwined teacher and student layers. Note that in Eqn.~(\ref{eq:fusing}), for simplicity we omit the adaption layers that are used for feature alignment. With the routing network $\mathcal{M}$, our ultimate goal is to obtain an isolated student model $S$ which performs on the task in interest as better as possible.

\subsection{The Policy Network}
We adopt a policy network to make decisions per sample on the data flow path through the routing network. Here we simply adopt a lightweight fully connected layer to implement the policy network. The input of the policy network is the concatenated features from the teacher and the student models. The outputs of the policy network are $N+1$ two-dimensional routing vectors, where $N+1$ denotes the number of branch points,
\textit{i.e.}, the number of candidate distillation spots. Each routing vector is a probability distribution, from which we stochastically draw a categorical value to make the decision on the data flow path for one branch point in the routing network. The sampling operation is not differentiable due to the discretization.
To enable the differentiability for the sampling operation, we utilize Gumbel-Softmax~\cite{GS} to implement the policy network. Formally, assume for the $i$-th branch point in the routing network, the corresponding routing vector is $\mathbf{a}^i_j=\{a^i_{1}, a^i_2\}$, where element $a^i_{1}$ stores the probability value that denotes how likely the teacher layers in the $i$-th block would be used to process the incoming data. During the forward propagation, the policy makes a discrete decision drawn from the categorical distribution based on the distribution:
\begin{equation}
\label{equ:receiver}
    \mathbf{w}^i=\text{one\_hot}\{\arg\max_k{(\log a^i_{k})}+\epsilon_k\}.
\end{equation}
Here $\mathbf{w}^i$ is a two-dimensional one-hot vector. ``one\_hot'' is the function returning a one-hot vector where only the specified element is 1 and all the others are 0. $\epsilon\in\mathbb{R}^{2}$ is a vector in which the elements are i.i.d samples drawn from the Gumbel distribution $(0, 1)$ to add a small amount of noise to avoid the argmax operation always selecting the element with the highest probability value.

To enable differentiability of the discrete sampling function, we use the Gumbel-Softmax trick to relax $\mathbf{w}^i$ during backward propagation as
\begin{equation}
\label{eq:gumbel-softmax}
    \mathbf{w}^i=\frac{\exp((\log \mathbf{a}^i + \epsilon)/\tau)}{\sum\nolimits_k\exp((\log a^i_{k}+\epsilon_k)/\tau)},
\end{equation}
where $\tau$ is the temperature that controls how sharp the distribution is after the approximation. Note that for each vector $\mathbf{w}^i$, as $w^i_1+w^i_2=1$, we simply use $w^i$ and $1-w^i$ to denote the routing decision on the teacher and the student layers, as shown in Eqn.~(\ref{eq:fusing}).

%\blue{As the policy network is implemented by only one fully connected layer, the extra GPU memory overhead is negligible compared to existing distillation methods.}

\subsection{Spot-adaptive Distillation}
The proposed spot-adaptive distillation is conducted by training the routing network and the policy network simultaneously. Training the proposed network is non-stationary from the perspective of both the policy network, and  the routing network, because the optimal routing strategy depends on the module parameters and vice versa. In this work, the multi-path routing network and the policy network are trained simultaneously in an end-to-end manner. The overall objective is
\begin{equation}
\label{eq:overall-loss}
    \mathcal{L}= \mathcal{L}_{student} + \beta_1\mathcal{L}_{KL} + \beta_2\mathcal{L}_{KD} + \beta_3\mathcal{L}_{routing},
\end{equation}
where $\mathcal{L}_{student}$ is the widely-used cross-entropy loss~\cite{de2005tutorial} between the targets and the predictions from only the student model. $\mathcal{L}_{KL}$ is the Kullback-Leibler divergence~\cite{hershey2007approximating} between the teacher and the student predictions, which is also the vanilla distillation loss proposed by Hinton~\textit{et al.}~\cite{Hinton2015DistillingTK}. $\mathcal{L}_{KD}$ is existing knowledge distillation loss imposed on the intermediate layers, such as the losses proposed by FitNets~\cite{Romero2015FitNetsHF}, Attention Transfer~\cite{Zagoruyko2017AT}, \textit{etc}. As the proposed method focuses on ``where to distill'' instead of ``what to distill'' that is mainly investigated by current literature, it can be combined with most existing distillation methods. $\mathcal{L}_{student}$, $\mathcal{L}_{KL}$ and $\mathcal{L}_{KD}$ are used to make the student model perform similarly with the teacher model. $\mathcal{L}_{routing}$ is the cross entropy loss between the targets and the predictions from routing network. $\beta_1$, $\beta_2$ and $\beta_3$ are three hyper-parameters for trading off these loss terms.

Note that during the whole training phase, the pre-trained parameters of the teacher model are kept fixed. The trainable parameters include only the parameters of the student models, the adaption layers and the policy netowrk. As the policy network and the adaption layers are involved in computing only $\mathcal{L}_{routing}$, their parameters are trained under the supervision of only $\mathcal{L}_{routing}$. The student network and the policy network forms a loop where the output of the student model goes into the policy network, and the output of the policy network goes into the student network again. To stabilize the training of the student network, we do not back-propagate the gradients from policy network to the student network anymore. At the early stage of the training process, as the teacher model is well pre-trained, samples are more likely to be passed to the teacher layers by the policy network. In this case, knowledge distillation happens at all candidate distillation spots. As the training proceeds, the student model gradually master the knowledge of the teacher to different degrees at different layers. In this situation, the policy network may plan a path per sample where both teacher layers and student layers are intertwined. Knowledge distillation is thus conducted adaptively at certain layers to push the optimal policy to involve only student layers. In the following section, we provide detailed optimization algorithm to more clearly describe the proposed distillation method.
\begin{algorithm}[t]
	\caption{SAKD Pseudocode, PyTorch-like}
	\footnotesize
	\begin{algorithmic}[1]
    \STATE \texttt{\anno{\# $S_i$: $i$-th block in the student model}}
    \STATE \texttt{\anno{\# $T_i$: $i$-th block in the teacher model}}
    \STATE \texttt{\anno{\# $P$: policy module}}
    \STATE \texttt{\anno{\# $H_{st}$, $H_{ts}$: adaption layer for adopting feature from student to teacher and from teacher to student, respectively}}
    \STATE
    \STATE \texttt{\anno{\# $\mathcal{L}_{CE}$: cross-entropy loss}}
    \STATE \texttt{\anno{\# $\mathcal{L}_{KL}$: Kullback-Leibler divergence}}
    \STATE \texttt{\anno{\# $\mathcal{L}_{KD}$: knowledge distillation loss}}
    \STATE \texttt{\anno{\# $\beta_1$,$\beta_2$,$\beta_3$: hyper-parameters for trading off loss terms}}
    \STATE
    \STATE $S$.train()
    \STATE $T$.eval()
    \STATE \texttt{\pink{for} $x$, target \pink{in} loader:
    \STATE \quad \anno{\# load a mini-batch ($x$, target) with $B$ samples}
    \STATE
    \STATE \quad \pink{with} torch.no\_grad():
    \STATE \quad \quad $logit_T$, $feat_T$ = $T$($x$)
    \STATE \quad $logit_S$, $feat_S$ = $S$($x$)
    \STATE
    \STATE \quad ft = torch.cat(($feat_T$[-1], $feat_S$[-1]), 1)
    \STATE \quad $w$ = $P$(ft)    \anno{\# shape: [$B$, $N+1$, $2$]}
    \STATE \quad $w$ = Gumbel-Softmax($w$)[:, :, 0] \anno{\# routing decisions}
    \STATE \quad $d$ = $w$.\pink{detach}() \anno{\# stop gradient}
    \STATE
    \STATE \quad \anno{\# loss for student}
    \STATE \quad $\mathcal{L}_s$ = $\mathcal{L}_{CE}$($logit_S$, target)
    \STATE \quad $\mathcal{L}_s$ += $\beta_1 \cdot$ $\mathcal{L}_{KL}$($logit_S$, $logit_T$) $\cdot$ $d$[:, -1]
    \STATE \quad $\mathcal{L}_{s}$ += $\beta_2 \cdot$ $\sum_{i}^N \mathcal{L}_{KD}$($feat_S$[i], $feat_T$[i]) $\cdot$ $d$[:, i]
    \STATE \quad
    \STATE \quad input\_s = $x$
    \STATE \quad input\_t = $x$
    \STATE \quad $S$.eval()
    \STATE \quad \anno{\# Multi-path Routing Network}
    \STATE \quad \pink{for} i \pink{in range}($1$, $N+1$):
    \STATE \quad \quad f\_t = $T_i$(input\_t)
    \STATE \quad \quad f\_s = $S_i$(input\_s)
    \STATE \quad \quad input\_s = f\_s $\cdot$ (1 - $w$[:, i]) + $H_{ts}$(f\_t) $\cdot$ $w$[:, i]
    \STATE \quad \quad input\_t = $H_{st}$(f\_s) $\cdot$ (1 - $w$[:, i]) + f\_t $\cdot$ $w$[:, i]
    \STATE \quad f\_t = $T_{N+1}$(input\_t)
    \STATE \quad f\_s = $S_{N+1}$(input\_s)
    \STATE
    \STATE \quad \anno{\# output from routing network}
    \STATE \quad out = f\_s $\cdot$ (1 - $w$[:, -1]) + f\_t $\cdot$ $w$[:, -1]
    \STATE \quad $S$.train()
    \STATE
    \STATE \quad \anno{\# loss for policy and adaption layers}
    \STATE \quad $\mathcal{L}_{routing}$ = $\beta_3$ $\cdot$ $\mathcal{L}_{CE}$(out, target)
    \STATE \quad $\mathcal{L}_{routing}$.\pink{backward()} \anno{\# back-propagate}
    \STATE \quad \pink{update}($P$, $H_{st}$, $H_{ts}$) \anno{\# SGD update}
    \STATE
    \STATE \quad $\mathcal{L}_s$.\pink{backward()} \anno{\# back-propagate}
    \STATE \quad \pink{update}(S) \anno{\# SGD update}}
	\end{algorithmic}
	\label{alg:sakd}
\end{algorithm}

\subsection{Optimization algorithm}
To make the proposed method clearer to readers, the pseudo-code of SAKD is provided in Algorithm~\ref{alg:sakd}.
Given two deep neural networks, a student $S$ and a teacher $T$. Let $x$ be the network input; we denote the lists of the intermediate representations from teacher and student models as $feat_T$ and $feat_S$, the final predictions as $logit_T$ and $logit_S$, respectively.
The input of the policy network $P$ is the concatenated features from the teacher and the student models, as seen in Line $\#20$.
The outputs of the policy model $P$ are $N + 1$ two-dimensional routing vectors, donates as $w$ (Line $\#21 \sim 22$), which are discrete decisions during the forward propagation and will be relaxed with Gumbel-softmax~\cite{GS} during backward propagation.
An apparent difficulty here is that the distillation loss~$\mathcal{L}_s$ for the student depend on the routing decision $w$, so it is problematic to optimize the student model together with the policy network.
We circumvent this difficulty by stop-gradient (Line $\#23$) operation. This means that $d$ is treated as a constant in the loss terms.
The complete objective for the student model is shown in Line $\#26 \sim 28$, which includes the cross-entropy loss, Kullback-Leibler divergence, and knowledge distillation loss.

After that, we begin the forward propagation of the multi-path routing network, which is composed of the teacher and the student networks, with their intermediate layers interconnected with each other.
We set the student to eval mode in advance in order to make the multi-path routing network work more steadily, and set the student back to train mode after gaining the final prediction of the routing network.
To align features between the teacher and the student, We adopt adaption layers (\textit{i.e.}, $H_{st}$ and $H_{ts}$), as shown in Line $\#37 \sim 38$.
The cross-entropy loss is used for optimizing the parameters of the policy module and adaption layers.

\section{Experiments}
\label{sec:experiments}
Here we first briefly describe the main experimental settings which are necessary to understand our experiments in this work. Then the benchmark comparisons are provided to demonstrate the superiority of the proposed method. Finally we conduct extensive ablation study to illustrate the effectiveness of the proposed method.

\subsection{Experimental Settings}
We compare the distillation performance with the top-1 and top-5 accuracy at the last epoch. Experiments are randomly repeated three times and the average results are provided.

\subsubsection{Datasets} Experiments are conducted on CIFAR-100~\cite{Krizhevsky2009LearningML}, tiny-ImageNet~\cite{Le2015TinyIV} and ImageNet~\cite{deng2009imagenet}. CIFAR-100 consists of 60,000 32$\times$32 colour images in 100 classes, with 600 images per class. There are 50,000 training images and 10,000 test images in the official split. The 100 classes in the CIFAR-100 are grouped into 20 superclasses. Tiny-ImageNet is a subset of ImageNet with 200 classes, where each image is down-sized to $64\times64$ pixels. Each class has 500 training images, 50 validation images, and 50 test images. For data augmentation, all images are normalized to zero mean and unit variance at each channel, we normalize each channel of all images to zero mean and unit variance for pre-processing. Images in CIFAR-100 are zero-padded with 4 pixels on every side, randomly cropped to 32$\times$32 pixels, then randomly horizontal flipped; and for tiny-ImageNet, we just employ random horizontal flipping. ImageNet consists of 1.2M training images and 50k validation images with 1,000 classes. Following the common practice, images in ImageNet are augmented with random cropping of size 224$\times$224 and horizontal flipping.

\subsubsection{Model Architectures} To validate the effectiveness of the proposed method, we adopt a variety of backbone architectures to implement the proposed model, including ResNet~\cite{He2016DeepRL}, WideResNet~(WRN)~\cite{BMVC2016_87}, Vgg~\cite{Simonyan2015VeryDC}, ShuffleNet V1~\cite{Zhang2018ShuffleNetAE}, MobieNet V2~\cite{Howard2017MobileNetsEC}. \blue{In this work, we adopt C-ResNet to denote the cifar-style ResNet with 3 groups of basic blocks, each with 16, 32, and 64 channels, respectively. and I-ResNet represents ImageNet-style ResNet with bottleneck blocks and more channels. More detailed configurations about the two types of networks can be found in~\cite{tian2019crd}}. WRN-d-w represents a WideResNet with depth $d$ and width factor $w$.

\subsubsection{Distillation Methods} The proposed adaptive distillation method can be combined with most existing distillation methods. We combine the proposed method with ten state-of-the-art distillation methods to demonstrate the superiority of the proposed method, including FitNets~\cite{Romero2015FitNetsHF}, AT~\cite{Zagoruyko2017AT}, SP~\cite{Tung2019SimilarityPreservingKD}, CC~\cite{Peng2019CorrelationCF}, VID~\cite{Ahn2019VariationalID}, RKD~\cite{Park2019RelationalKD}, PKT~\cite{Passalis2018LearningDR}, FT~\cite{Kim2018ParaphrasingCN}, NST~\cite{Huang2017LikeWY}, CRD~\cite{tian2019crd}. For all methods, we combine their objectives with the vanilla KD~\cite{Hinton2015DistillingTK} objective, the KL divergence between the softened predictions from teacher and the student models, to boost their performance. Thus all the methods involve at least two distillation spots, turning into a multi-spot version no matter they are originally one-spot or multi-spot distillation methods.

\subsubsection{Distillation Spots}
In this work, we combine the proposed spot-adaptive KD (SAKD) with ten state-of-the-art knowledge distillation methods, including FitNets~\cite{Romero2015FitNetsHF}, AT~\cite{Zagoruyko2017AT}, SP~\cite{Tung2019SimilarityPreservingKD}, CC~\cite{Peng2019CorrelationCF}, VID~\cite{Ahn2019VariationalID}, RKD~\cite{Park2019RelationalKD}, PKT~\cite{Passalis2018LearningDR}, FT~\cite{Kim2018ParaphrasingCN}, NST~\cite{Huang2017LikeWY}, CRD~\cite{tian2019crd}.
Additionally, we bond these objectives with the vanilla KD~\cite{Hinton2015DistillingTK} objective.
The above distillation methods determine different distillation spots before training and remain the same throughout the whole distillation process.
%For a clear understanding of distillation spots with various methods, we visualize all candidate distillation spots in Fig.~\ref{fig:candidate-distillation-spots}.
We assume that the network consists of $N$ blocks and one linear function.
If one method adopts the output from Block $i$ ($1 \leq i \leq N+1$) learned by the teacher as knowledge for training the student, its distillation spot is referred to as $i$.
The detailed experimental setups about distillation spots are shown in Table~\ref{tab:distillation-spots}.
%
%\begin{figure}[t]
%    \centering
%    \includegraphics[scale=0.408]{./fig/distillation-spots.pdf}
%    \caption{ An illustration of general candidate distillation spots. }
%    \label{fig:candidate-distillation-spots}
%\end{figure}
%\vspace{1.0pt}

\begin{table}[t]
    \begin{center}
    \resizebox{0.48\textwidth}{!}{
    \begin{tabular}{c|cccc}
    \toprule
        Method & Fitnets~\cite{Romero2015FitNetsHF} & AT~\cite{Zagoruyko2017AT} & SP~\cite{Tung2019SimilarityPreservingKD} & CC~\cite{Peng2019CorrelationCF} \\
        DS. & $3$ & $2, 3, \dots, N-1$ & $N-1$ & $N$  \\
        \midrule
        Method & VID~\cite{Ahn2019VariationalID} & RKD~\cite{Park2019RelationalKD} & PKT~\cite{Passalis2018LearningDR} & FT~\cite{Kim2018ParaphrasingCN} \\
        DS. & $2, 3, \dots, N-1$ & $N$ & $N$ & $N-1$ \\
        \midrule
        Method   & NST~\cite{Huang2017LikeWY} & CRD~\cite{tian2019crd} & KD~\cite{Hinton2015DistillingTK}  & \\
        % \midrule
        DS.  & $2, 3, \dots, N-1$ & $N$ & $N+1$  & \\
        \bottomrule
    \end{tabular}
    }
    \end{center}
    \caption{Detailed setting of distillation spots on various distillation methods. ``DS.'' denotes distillation spots.}
    \label{tab:distillation-spots}
\end{table}

\begin{table}[t]
    \begin{center}
    \resizebox{0.48\textwidth}{!}{
    \begin{tabular}{c|ccccc}
    \toprule
        Method & Fitnets~\cite{Romero2015FitNetsHF} & AT~\cite{Zagoruyko2017AT} & SP~\cite{Tung2019SimilarityPreservingKD} & CC~\cite{Peng2019CorrelationCF} & VID~\cite{Ahn2019VariationalID}  \\
        %\midrule
        $\beta_2$ & 1.0 & 1,000 & 3,000 & 0.02 & 1.0  \\
        \midrule
        Method & RKD~\cite{Park2019RelationalKD} & PKT~\cite{Passalis2018LearningDR} & FT~\cite{Kim2018ParaphrasingCN} & NST~\cite{Huang2017LikeWY} & CRD~\cite{tian2019crd}  \\
        % \midrule
        $\beta_2$ & 1.0 & 30,000 & 200 & 50  & 0.8 \\
        \bottomrule
    \end{tabular}
    }
    \end{center}
    \caption{Setting of the distillation loss factor ($\beta_2$).}
    \label{tab:hyper-parameter}
\end{table}

\begin{table*}[!ht]
\small
\begin{center}
\resizebox{\textwidth}{!}{
\begin{tabular}{c|ccc|ccc|ccc|ccc|cccccc}
  \toprule
  \multicolumn{1}{c}{}&\multicolumn{3}{c}{\textbf{C-ResNet56$\rightarrow$C-ResNet20}} &\multicolumn{3}{c}{\textbf{C-ResNet110$\rightarrow$C-ResNet32}} &\multicolumn{3}{c}{\textbf{C-ResNet32$\times$4$\rightarrow$C-ResNet8$\times$4}} &\multicolumn{3}{c}{\textbf{Vgg13$\rightarrow$Vgg8}}
  &\multicolumn{3}{c}{\textbf{WRN\_40\_2$\rightarrow$WRN\_16\_2}}\\
  %\cmidrule(lr){2-4} \cmidrule(lr){5-7} \cmidrule(lr){8-10}
  \midrule
   &Stan. &Adap. &$\Delta$ &Stan. &Adap. &$\Delta$ &Stan. &Adap. &$\Delta$ &Stan. &Adap. &$\Delta$ &Stan. &Adap. &$\Delta$ \\
  \midrule
  \textbf{T} & 72.34 & -- &--  & 74.31 & -- &--  &79.42 &-- & -- & 74.64 & -- &--  & 75.61 & -- &--  \\
  \textbf{S} & 69.06 & -- &--  & 71.14 & -- &--  &72.50 &-- & -- &70.36 &-- &-- &73.26 &-- &--\\
  {KD}      & 70.66  & -- &--  & 73.08 & -- &--  &73.33    &-- &-- &72.98 &-- &-- &74.92 &-- &--\\
  \midrule
                   %---------------------------------------------------------------------------------
  {Fitnets} & 71.05 & 71.40 &\green{+0.35}
                   & 73.19  & 73.58  &\green{+0.39}
                   & 74.66  & 74.95  &\green{+0.29}
                   & 73.22  & 73.54  &\green{+0.32}
                   & 75.12  & 75.29  &\green{+0.17}\\
  {AT}      & 70.99  & 71.41  &\green{+0.42}
                   & 73.16  & 73.47  &\green{+0.31}
                   & 74.53  & 75.53  &\green{+1.00}
                   & 73.48  & 73.81  &\green{+0.33}
                   & 75.32  & 75.15  &\red{-0.17}\\
  {SP}     & 70.65   & 71.22  & \green{+0.57}
                   & 73.03   & 73.26  &\green{+0.23}
                   & 74.02   & 74.64  &\green{+0.60}
                   & 73.49   & 73.61  &\green{+0.12}
                   & 74.98   & 75.06  &\green{+0.08}\\
  {CC}      & 71.03   & 71.48  &\green{+0.45}
                   & 73.07  & 73.42  &\green{+0.35}
                   & 74.21   & 74.64  &\green{+0.43}
                   & 73.04   & 73.26 &\green{+0.22}
                   & 75.09   & 75.41 &\green{+0.32}\\
  {VID}     & 71.06   & 71.34  & \green{+0.28}
                   & 73.31   & 73.71  &\green{+0.40}
                   & 74.56   & 75.26  &\green{+0.70}
                   & 73.19  & 74.12   &\green{+0.93}
                   & 75.14  & 75.27  &\green{+0.13}\\
  {RKD}    & 71.07   & 71.33  & \green{+0.26}
                   & 72.87   & 73.17  &\green{+0.30}
                   & 73.79   & 74.44  &\green{+0.65}
                   & 72.97  & 73.44  &\green{+0.47}
                   & 74.89  & 75.21  &\green{+0.32}\\
  {PKT}     & 70.72   & 71.56  &\green{+0.84}
                   & 73.61   & 73.92  &\green{+0.31}
                   & 74.23   & 74.49  & \green{+0.26}
                   & 73.25   & 73.53  &\green{+0.28}
                   & 75.33   & 75.49  &\green{+0.16}\\
  {FT}      & 71.15   & 71.37  & \green{+0.22}
                   & 73.44   & 73.65  &\green{+0.21}
                   & 74.62   & 75.19  &\green{+0.57}
                   & 73.44  & 73.46   &\green{+0.02}
                   & 75.15  & 75.30   &\green{+0.15}\\
  {NST}     & 70.68   & 71.07  & \green{+0.39}
                   & 72.91   & 73.50  &\green{+0.59}
                   & 74.28  & 75.16 & \green{+0.88}
                   & 73.33  & 75.44 &\green{+0.31}
                   & 74.67  & 75.27 &\green{+0.60}\\
  {CRD}     & 71.46   & 71.78  & \green{+0.32}
                   & 73.58   &74.25  &\green{+0.67}
                   & 75.46  & 75.81 & \green{+0.25}
                   & 74.29  & 74.53 & \green{+0.24}
                   & 75.64  & 76.03 & \green{+0.39}\\
  \bottomrule
\end{tabular}
}
\end{center}
\caption{Top-1 accuracy of homogeneous distillation on CIFAR-100~(in $\%$). Experiments are repeated for three times and the average results are provided. ``Stan.'' denotes the standard distillation strategy in existing methods. ``Adap.'' denotes the adaptive distillation strategy proposed in this work.  $\Delta$: accuracy difference between ``Adap.'' and ``Stan.''. Numbers in~\green{green}~(\red{red}) denote that ``Adap.'' \green{outperforms}~(\red{underperforms}) ``Stan.''. T and S represent the teacher and the student models, respectively.}
\label{table:homegeneous}
\end{table*}

\subsubsection{Implementation Details} For fair comparisons, we keep the implementation settings of all the experiments the same. Specifically, following CRD~\cite{tian2019crd}, we set the batch size to 64, the number of total training epochs to 240. Most  experiments employ SGD as the optimizer for training the student model. The initial learning rate is set to $0.05$ for most backbone architectures, except MobineNet V2 and ShuffleNet V1 where the initial learning rate is set to $0.01$. Weight decay is set to be $0.0005$. The learning rate decays by a factor of $0.1$ at the $150$-th, the $180$-th, and the $210$-th epochs. The temperature value for softening the predicted distributions is set to 4. $\tau$ in Gumbel-Softmax is initially set to 5 and decays gradually during training so the network can explore freely in the early stage and exploit the converged distillation policy in the later stage. For simplicity, the hyper-parameters $\beta_1$ and $\beta_3$ in Eqn.~\ref{eq:overall-loss} are set to 1. $\beta_2$ is set based on the distillation methods. We adopt corresponding hyper-parameters in CRD~\cite{tian2019crd} to set $\beta_2$ for most distillation methods, except FitNets where $\beta_2$ is set to 1 instead of $1,000$ for more stable training. Detailed settings of $\beta_{2}$ are shown in Table~\ref{tab:hyper-parameter}.

\begin{table*}[t]
\small
\begin{center}
\resizebox{\textwidth}{!}{
\begin{tabular}{c|ccc|ccc|ccc|ccc|cccccc}
  \toprule
  \multicolumn{1}{c}{}&\multicolumn{3}{c}{\textbf{Vgg13$\rightarrow$MobileNetV2}} &\multicolumn{3}{c}{\textbf{I-ResNet50$\rightarrow$MobileNetV2}} &\multicolumn{3}{c}{\textbf{I-ResNet50$\rightarrow$Vgg8}} &\multicolumn{3}{c}{\textbf{C-ResNet32$\times$4$\rightarrow$C-ResNet32}}
  &\multicolumn{3}{c}{\textbf{WRN\_40\_2$\rightarrow$ShuffleNetV1}}\\
  %\cmidrule(lr){2-4} \cmidrule(lr){5-7} \cmidrule(lr){8-10}
  \midrule
   &Stan. &Adap. &$\Delta$ &Stan. &Adap. &$\Delta$ &Stan. &Adap. &$\Delta$ &Stan. &Adap. &$\Delta$ &Stan. &Adap. &$\Delta$ \\
  \midrule
  \textbf{T} & 74.64 & -- &--  & 79.34 & -- &--  &79.34 &-- & -- & 74.92 & -- &--  & 75.61 & -- &--  \\
  \textbf{S} & 64.60 & -- &--  & 64.60 & -- &--  &70.36 &-- & -- &71.14 &-- &-- &70.50 &-- &--\\
  {KD}      & 67.37  & -- &--  & 67.35 & -- &--  &73.81    &-- &-- &72.98 &-- &-- &74.83 &-- &--\\
  \midrule
                   %---------------------------------------------------------------------------------
  {Fitnets} & 72.47 & 73.11 &\green{+0.64}
                   & 72.69  & 73.03  &\green{+0.34}
                   & 73.24  & 73.75  &\green{+0.51}
                   & 72.35  & 72.63  &\green{+0.28}
                   & 75.67  & 75.93  &\green{+0.26}\\
  {AT}      & 72.00  & 72.60  &\green{+0.60}
                   & 71.54 & 72.11  &\green{+0.57}
                   & 74.01  & 73.82  &\red{-0.19}
                   & 72.67  & 72.86  &\green{+0.19}
                   & 76.24  & 76.55  &\green{+0.29}\\
  {SP}     & 73.04   & 73.31  & \green{+0.27}
                   & 73.17   & 73.42  &\green{+0.25}
                   & 73.52   & 74.23  &\green{+0.71}
                   & 71.79   & 72.65  &\green{+0.86}
                   & 76.29   & 76.54  &\green{+0.25}\\
  {CC}      & 72.41   & 72.93  &\green{+0.52}
                   & 72.61  & 72.73  &\green{+0.32}
                   & 73.48   & 73.54  &\green{+0.06}
                   & 72.37   & 72.71 &\green{+0.34}
                   & 75.24   & 75.77 &\green{+0.53}\\
  {VID}     & 72.10   & 72.70  & \green{+0.60}
                   & 72.58   & 73.08  &\green{+0.50}
                   & 73.46   & 73.92  &\green{+0.46}
                   & 72.29  & 72.85   &\green{+0.56}
                   & 75.88  & 76.21  &\green{+0.33}\\
  {RKD}    & 72.58   & 72.87  & \green{+0.29}
                   & 72.90   & 73.86  &\green{+0.96}
                   & 73.51   & 73.85  &\green{+0.34}
                   & 71.35  & 72.17  &\green{+0.82}
                   & 75.66  & 75.88  &\green{+0.22}\\
  {PKT}     & 72.76   & 73.23  &\green{+0.47}
                   & 73.01   & 73.41  &\green{+0.40}
                   & 73.61   & 73.90  & \green{+0.29}
                   & 72.04   & 72.75  &\green{+0.71}
                   & 75.66   & 76.05  &\green{+0.39}\\
  {FT}      & 72.02   & 72.30  & \green{+0.28}
                   & 72.85   & 73.11  &\green{+0.26}
                   & 72.98   & 73.54  &\green{+0.56}
                   & 72.42  & 73.46   &\green{+1.02}
                   & 72.96  & 73.60   &\green{+0.64}\\
  {NST}     & 72.31   & 72.66  & \green{+0.35}
                   & 73.01   & 73.41  &\green{+0.40}
                   & 71.74  & 72.00 & \green{+0.26}
                   & 72.34  & 72.69 &\green{+0.35}
                   & 76.45  & 75.96 &\red{-0.49}\\
  {CRD}     & 73.56   & 73.81  & \green{+0.25}
                   & 73.76   &74.29  &\green{+0.53}
                   & 74.58  & 74.81 & \green{+0.23}
                   & 72.99  & 73.21 & \green{+0.22}
                   & 76.03  & 76.26 & \green{+0.23}\\
  \bottomrule
\end{tabular}
}
\end{center}
\caption{Top-1 accuracy of heterogeneous distillation on CIFAR-100~(in $\%$). Experiments are repeated for three times and the average results are provided. }
\label{table:heterogeneous}
\end{table*}

\begin{table*}[t]
\small
\begin{center}
\resizebox{\textwidth}{!}{
\begin{tabular}{c|ccc|ccc|ccc|ccc|cccccc}
  \toprule
  \multicolumn{1}{c}{}&\multicolumn{3}{c}{\textbf{C-ResNet56$\rightarrow$C-ResNet20}} &\multicolumn{3}{c}{\textbf{C-ResNet110$\rightarrow$C-ResNet20}} &\multicolumn{3}{c}{\textbf{Vgg13 $\rightarrow$ Vgg8}} &\multicolumn{3}{c}{\textbf{WRN\_40\_2 $\rightarrow$ WRN\_16\_2}}
  &\multicolumn{3}{c}{\textbf{Vgg13 $\rightarrow$ MobileNetV2}}\\
  %\cmidrule(lr){2-4} \cmidrule(lr){5-7} \cmidrule(lr){8-10}
  \midrule
   &Stan. &Adap. &$\Delta$ &Stan. &Adap. &$\Delta$ &Stan. &Adap. &$\Delta$ &Stan. &Adap. &$\Delta$ &Stan. &Adap. &$\Delta$ \\
  \midrule
  \textbf{T} & 58.34 & -- &--  & 58.46 & -- &--  &60.09 &-- & -- & 61.26 & -- &--  & 60.09 & -- &--  \\
  \textbf{S} & 52.66 & -- &--  & 51.89 & -- &--  &56.03 &-- & -- & 57.17 &-- &-- &57.73 &-- &--\\
  {KD}      & 53.04  & -- &--  & 53.40 & -- &--  &57.33    &-- &-- & 59.16 &-- &-- &60.02 &-- &--\\
                   %---------------------------------------------------------------------------------
  \midrule
  {Fitnets}      & 54.43   & 54.53  &{\green{+0.10}}
                   & 54.04   & 54.25  &\green{+0.21}
                   & 58.33   & 59.10  &\green{+0.77}
                   & 58.88   & 59.33  &\green{+0.45}
                   & 61.37   & 61.83  &\green{+0.46}
                   \\
  {AT}     & 54.39   & 54.88  & \green{+0.49}
                   & 54.57   & 54.71  & \green{+0.14}
                   & 58.85   & 59.31  & \green{+0.46}
                   & 59.39   & 59.65  & \green{+0.26}
                   & 60.84   & 61.34  & \green{+0.50}
                   \\
  {FT}      & 53.90   & 54.32  &{\green{+0.42}}
                   & 54.46   & 55.10  &\green{+0.64}
                   & 58.87   & 59.21  &\green{+0.34}
                   & 58.85   & 58.90  &\green{+0.05}
                   & 61.78   & 61.96  &\green{+0.18}
                   \\
  {PKT}     & 54.29   & 54.50  & \green{+0.21}
                   & 54.70   & 55.01  & \green{+0.31}
                   & 58.87   & 59.13  & \green{+0.26}
                   & 59.19   & 59.59  & \green{+0.40}
                   & 61.90   & 62.14  & \green{+0.24}
                   \\
  {SP}    & 54.23   & 54.39  & \green{+0.16}
                   & 54.38   & 54.52  & \green{+0.14}
                   & 58.78   & 59.26  & \green{+0.48}
                   & 57.63   & 58.26  & \green{+0.13}
                   & 61.90   & 62.29  & \green{+0.39}
                   \\
  {VID}     & 53.89   & 53.95  &\green{+0.06}
                   & 53.94   & 54.28  & \green{+0.34}
                   & 58.55   & 58.80  & \green{+0.25}
                   & 58.78   & 58.99  & \green{+0.21}
                   & 60.84   & 61.23  & \green{+0.39}
                   \\
  {CC}      & 54.22   & 54.83  &\green{+0.63}
                   & 54.26   & 54.35  &\green{+0.09}
                   & 58.18   & 58.67  &\green{+0.49}
                   & 58.83   & 59.08  &\green{+0.25}
                   & 61.32   & 61.82  &\green{+0.50}
                   \\
  {RKD}     & 53.95   & 54.05  &\green{+0.10}
                   & 53.88   & 54.09  &\green{+0.21}
                   & 58.58  & 58.62 & \green{+0.04}
                   & 59.31  & 59.32 & \green{+0.01}
                   & 61.19  & 61.58 & \green{+0.39}
                   \\
  {NST}    & 53.66   & 54.27  & \green{+0.61}
                   & 53.82   & 54.01  & \green{+0.19}
                   & 58.85   & 59.53  & \green{+0.68}
                   & 59.07   & 59.20  & \green{+0.13}
                   & 60.59   & 60.64  & \green{+0.05}
                   \\
  {CRD}     & 55.04   & 55.06  &\green{+0.02}
                   & 54.69   &55.28  &\green{+0.59}
                   & 58.88  & 59.38 & \green{+0.50}
                   & 59.42 & 59.87 & \green{+0.45}
                   & 61.63 & 61.89 & \green{+0.26}\\
  \bottomrule
\end{tabular}
}
\end{center}
\caption{Top-1 accuracy of adaptive distillation on tiny-ImageNet~(in $\%$). Both homogeneous and heterogeneous distillation are included here.}
\label{table:im-home-heter}
\end{table*}

\subsection{Benchmark Comparisons}
%The code and models will be made publicly available soon for reproducing the provided results in the paper.
\subsubsection{Homogeneous Distillation}
In this section, we evaluate the proposed method under homogeneous distillation. To this end, we adopt various teacher-student combinations, where the teacher models and the student models are in the same-style architectures, including C-ResNet56 $\rightarrow$ C-ResNet20, C-ResNet110 $\rightarrow$ C-ResNet32, C-ResNet32x4 $\rightarrow$ C-ResNet8x4, Vgg13 $\rightarrow$ Vgg8, WRN\_40\_2 $\rightarrow$ WRN\_16\_2. In our experiments, the candidate distillation spots include the softmax layer for $\mathcal{L}_{KL}$, and some intermediate layers for $\mathcal{L}_{KD}$.  Note that all competitors are combined with KD~\cite{Hinton2015DistillingTK} to enhance their performance in the standard distillation scheme, and thus the softmax layer is always a candidate distillation spot in the proposed adaptive scheme. However, for intermediate layers, different distillers conduct knowledge distillation at different number of layers~(one-spot distillation or multi-spot distillation). The proposed adaptive distillation strategy only determines where or not to conduct distillation at these distillation spots. It does not add any other candidate distillation spots to the standard distillation methods.

Experimental results on CIFAR-100 are shown in Table~\ref{table:homegeneous}, and results on tiny-ImageNet are shown in Table~\ref{table:im-home-heter}. From these results, it can be seen that the proposed adaptive distillation strategy consistently outperforms the standard distillation strategy with nearly all distillers, under all teacher-student architectures, and on both CIFAR-100 and tiny-ImageNet datasets. For example, when combined with PKT~\cite{Passalis2018LearningDR} for distilling knowledge from C-ResNet56 to C-ResNet20, the proposed adaptive distillation strategy improves the accuracy by $0.84\%$~(from $70.72\%$ to $71.56\%$) and $0.21\%$~(from $54.29\%$ to $54.50\%$), respectively. The standard distillation strategy imposes distillation constraints on all distillation spots and throughout the whole distillation process, which leads to over-regularization on the student and thus the performance degradation. Similar results are also observed on other distillation methods and student-teacher combinations. The highly consistent results demonstrate the effectiveness of the proposed method for improving performance of existing distillation methods.

\subsubsection{Heterogeneous Distillation}
The proposed adaptive distillation strategy is naturally compatible with homogeneous distillation. However, it seems not so compatible with heterogeneous distillation. In this section, we evaluate the proposed method under heterogeneous distillation settings. On CIFAR-100, the adopted teacher-student combinations include Vgg13 $\rightarrow$ MobileNetV2, I-ResNet50 $\rightarrow$ MobileNetV2, I-ResNet50 $\rightarrow$ Vgg8, C-ResNet32$\times4$ $\rightarrow$ C-ResNet32 and WRN\_40\_2 $\rightarrow$ ShuffleNetV1. On tiny-ImageNet, we provide results of only Vgg13 $\rightarrow$ MobileNetV2 for space consideration, as other teacher-student combinations produce similar results. Similar to homogeneous distillation, the candidate distillation spots include the softmax layer for $\mathcal{L}_{KL}$, and intermediate layers for $\mathcal{L}_{KD}$.  The softmax layer is always a candidate distillation spot in the proposed adaptive scheme.

Experimental results on CIFAR-100 are provided in Table~\ref{table:heterogeneous}, and the results on tiny-ImageNet are provided in Table~\ref{table:im-home-heter}~(Vgg13 $\rightarrow$ MobileNetV2). The results show that the proposed adaptive distillation strategy also consistently improves performance of existing distillation methods under the heterogeneous distillation settings. For example, when combined with VID~\cite{Ahn2019VariationalID} for distilling knowledge from Vgg13 to MobileNetV2, the proposed adaptive distillation strategy improves the accuracy by $0.60\%$~(from $72.10\%$ to $72.70\%$) and $0.39\%$~(from $60.84\%$ to $61.23\%$), respectively. The highly consistent results again demonstrate the effectiveness and universality of the proposed spot-adaptive distillation strategy across various experimental settings.

\begin{table}[t]
    \begin{center}
    \resizebox{0.48\textwidth}{!}{
    \begin{tabular}{c|c|cccc}
    \toprule
        &Method & AT & SP & CC & CRD  \\
        \hline
        \multirow{3}{*}{top$@$1}
        &Standard &70.70 &70.62 &69.96 &71.38\\
        &Adaptive &70.94 &71.17 &70.72 &71.63\\
        &$\Delta$ &\green{+0.24} &\green{+0.55} &\green{+0.76} &\green{+0.25}\\
        \hline
        \multirow{3}{*}{top$@$5}
        &Standard &90.00 &89.80 &89.17 &90.49\\
        &Adaptive &90.33 &90.22 &90.28 &90.85\\
        &$\Delta$ &\green{+0.33} &\green{+0.42} &\green{+1.11} &\green{+0.36}\\
        \bottomrule
    \end{tabular}
    }
    \end{center}
    \caption{Top@1 and top@5 accuracy results on ImageNet. Here I-ResNet34 is used as the teacher network, and I-ResNet18 as the student.}
    \label{tab:imagenet}
\end{table}

\subsubsection{Results on ImageNet} To validate the scalability of the proposed method on large-scale datasets, here we evaluate the proposed method on ImageNet. For a fair comparison with existing methods~\cite{tian2019crd,Zagoruyko2017AT} that have been tested on ImageNet, here we follow their experimental settings, adpoting ResNet-34 as the teacher and ResNet-18 as the student. Experimental results are provided in Table~\ref{tab:imagenet}. It can be seen that the proposed adaptive distillation strategy also consistently improves both top@1 and top@5 accuracy of existing distillers on ImageNet. For examples, with the proposed method, CC~\cite{Peng2019CorrelationCF} improves its top@1 accuracy from $69.96\%$ to $70.72\%$, and top@5 from $89.17\%$ to $90.28\%$. These results verify the scalability of the proposed method to larger datasets.

\begin{table*}[t]
\small
\begin{center}

\resizebox{\textwidth}{!}{
\begin{tabular}{c|ccccc|ccccc|ccccc}
  \toprule
   &\textbf{Adaptive} &\textbf{Always} & \textbf{Anti} &\textbf{Rand} &\textbf{No} &\textbf{Adaptive} &\textbf{Always} & \textbf{Anti} &\textbf{Rand} &\textbf{No} &\textbf{Adaptive} &\textbf{Always} & \textbf{Anti} &\textbf{Rand} &\textbf{No}\\
   \midrule
     \multicolumn{1}{c}{\textbf{CIFAR100}}&\multicolumn{5}{c}{{C-ResNet32$\times$4$\rightarrow$C-ResNet8$\times$4}} &\multicolumn{5}{c}{{Vgg13$\rightarrow$Vgg8}} &\multicolumn{5}{c}{{WRN\_40\_2$\rightarrow$WRN\_16\_2}}\\
   \midrule
                   %---------------------------------------------------------------------------------
  {Fitnets}  & \textbf{74.95}  & 74.66 & 73.22 & 74.23 & 72.50
            & \textbf{73.54}  & 73.22& 71.37 & 72.82 & 70.36
            & \textbf{75.29}  & 75.12 & 73.56 & 75.17 & 73.26\\
  {AT}
            & \textbf{75.53}  & 74.53 & 73.24 & 75.04 & 72.50
            & \textbf{73.81}  & 73.48 & 71.43 & 73.59 & 70.36
            & 75.15  & \textbf{75.32} & 72.47 & 74.83 &73.26\\
  {SP}
            & \textbf{74.64}  & 74.02 & 71.89 & 73.57 & 72.50
            & \textbf{73.61}  & 73.49 & 69.91 & 72.54 & 70.36
            & \textbf{75.06}  & 74.98 & 70.32 & 71.25 & 73.26\\
  {VID}
            & \textbf{75.26}  & 74.56 & 73.02 & 74.52 &72.50
            & \textbf{74.12}  & 73.19 & 71.16 & 72.87 &70.36
            & \textbf{75.27}  & 75.14 & 73.44 & 74.76 & 73.26\\
  {CRD}
            & \textbf{75.81}  & 75.46 & 72.38 & 75.42 & 72.50
            & \textbf{74.53}  & 74.29 & 70.38 & 73.41 & 70.36
            & \textbf{76.03}  & 75.64 & 72.84 & 75.41 & 73.26\\
  \midrule
     \multicolumn{1}{c}{\textbf{CIFAR100}}&\multicolumn{5}{c}{Vgg13$\rightarrow$MobileNetV2} &\multicolumn{5}{c}{{I-ResNet50$\rightarrow$MobineNetV2}} &\multicolumn{5}{c}{WRN\_40\_2$\rightarrow$ShuffleNetV1}\\
   \midrule
   {Fitnets} & \textbf{73.11}  & 72.47 & 69.98  & 72.01 & 64.60
            & \textbf{73.03}  & 72.69 & 69.00  & 72.61  & 64.60
            & \textbf{75.93}  & 75.67 & 73.00  & 75.37  & 73.26\\
  {AT}      & \textbf{72.60}  & 72.00 & 67.39  & 71.33  & 64.60
            & \textbf{72.11}  & 71.54 & 68.85  & 70.28  & 64.60
            & \textbf{76.55}  & 76.24 & 73.77  & 75.74  & 73.26\\
  {SP}      & \textbf{73.31}  & 73.04 & 65.41  & 72.09  & 64.60
            & \textbf{73.42}  & 73.17 & 64.61  & 72.62  & 64.60
            & \textbf{76.54}  & 76.29 & 70.51  & 75.21  & 73.26\\
  {VID}     & \textbf{72.70}  & 72.10 & 69.72  & 72.13  & 64.60
            & \textbf{73.08}  & 72.58 & 68.33  & 72.61  & 64.60
            & \textbf{76.21}  & 75.88 & 72.24  & 75.67  & 73.26\\
  {CRD}     & \textbf{73.81}  & 73.56 & 68.08  & 73.25  & 64.60
            & \textbf{74.29}  & 73.76 & 68.78  & 73.48  & 64.60
            & \textbf{76.26}  & 76.03 & 71.50  & 75.75  & 73.26\\
    \midrule
     \multicolumn{1}{c}{\textbf{tiny-ImageNet}}&\multicolumn{5}{c}{Vgg13$\rightarrow$Vgg8} &\multicolumn{5}{c}{{WRN\_40\_2$\rightarrow$WRN\_16\_2}} &\multicolumn{5}{c}{Vgg13$\rightarrow$MobileNetV2}\\
     \midrule
     {SP}   & \textbf{59.26}  & 58.78 & 55.65  & 57.92  & 56.03
            & \textbf{58.26}  & 57.63 & 56.65  & 54.25  & 57.17
            & \textbf{62.29}  & 61.90 & 59.24  & 60.67  & 57.73\\
  {CRD}     & \textbf{59.38}  & 58.88 & 54.97  & 58.14  & 56.03
            & \textbf{59.87}  & 59.42 & 56.09  & 59.29  & 57.17
            & \textbf{61.89}  & 61.63 & 58.02  & 61.11  & 57.73\\
  \bottomrule
\end{tabular}
}

\end{center}
\caption{Ablation study of the proposed spot-adaptive distillation strategy on CIFAR-100 and tiny-ImageNet~(in $\%$).}
\label{table:distillation-ablation}
\end{table*}

\subsection{Ablation Study}
We perform ablation study in this section to demonstrate the effectiveness of the proposed method. Specifically, we answer the following questions with carefully designed experiments.
%\vspace{5.5pt}

\noindent\textit{1) Does the policy network make useful decisions?}

Here we validate usefulness of the decision made by the policy network. To this end, we introduce four baseline distillation strategies: \textit{always-distillation}, \textit{rand-distillation}, \textit{anti-distillation} and \textit{no-distillation}.
Always-distillation is actually the standard distillation strategy where distillation is always conducted at every distillation spot.
Rand-distillation, as its name implies, randomly decides on whether or not to make the distillation at candidate distillation spots.
Anti-distillation adopts an opposite distillation strategy to the proposed adaptive distillation: if adaptive distillation distills at a certain spot, anti-distillation does not distill; otherwise, it distills at this spot. No-distillation means that the student is trivially trained without any distillation.
Experiments are conducted on CIFAR-100 and tiny-ImageNet with different KD methods and network pairs under both homogeneous and heterogeneous distillation. The results are provided in Table~\ref{table:distillation-ablation}.  It can be seen that the proposed adaptive distillation consistently outperforms other baselines, including the competitive always-distillation. Although the improvement
is sometimes marginal for some distillation methods, the consistent improvement with almost all the distillation verifies that the proposed policy network indeed makes useful routing decisions for distillation. Furthermore, anti-distillation usually produces much worse performance than adaptive-, always- and rand-distillation, sometimes even worse than no-distillation~(\textit{e.g.}, WRN\_40\_2$\rightarrow$ShuffleNetV1 on CIFAR100). These results indicate that conducting distillation at inappropriate spots can be harmful for training the student.

\iffalse
To make the conclusion more convincing, we provide more experimental results of these distillation strategies on CIFAR-100.
As shown in Table~\ref{table:distillation-ablation}, both homogeneous and heterogeneous distillations are provided.
We selected five representative distillation methods to cooperate with SAKD: Fitnets~\cite{Romero2015FitNetsHF}, AT~\cite{Zagoruyko2017AT}, SP~\cite{Tung2019SimilarityPreservingKD}, VID~\cite{Ahn2019VariationalID}, CRD~\cite{tian2019crd}.
These five knowledge distillation methods cover all conditions of distillation spots, as detailed in Table~\ref{tab:distillation-spots}.
Compared to anti-distillation and random-distillation, adaptive distillation always achieves the best performance, indicating that adaptive distillation strategy is able to make the right choices effectively.
Besides, we observe that training the student model with the anti-distillation sometimes gets lower accuracy than no-distillation, which means the anti-distillation strategy may be harmful to the student model.

All these results convincingly evidence the effectiveness of the proposed adaptive distillation strategy.
\fi

\begin{figure}[t]
    \centering
    \includegraphics[scale=0.293]{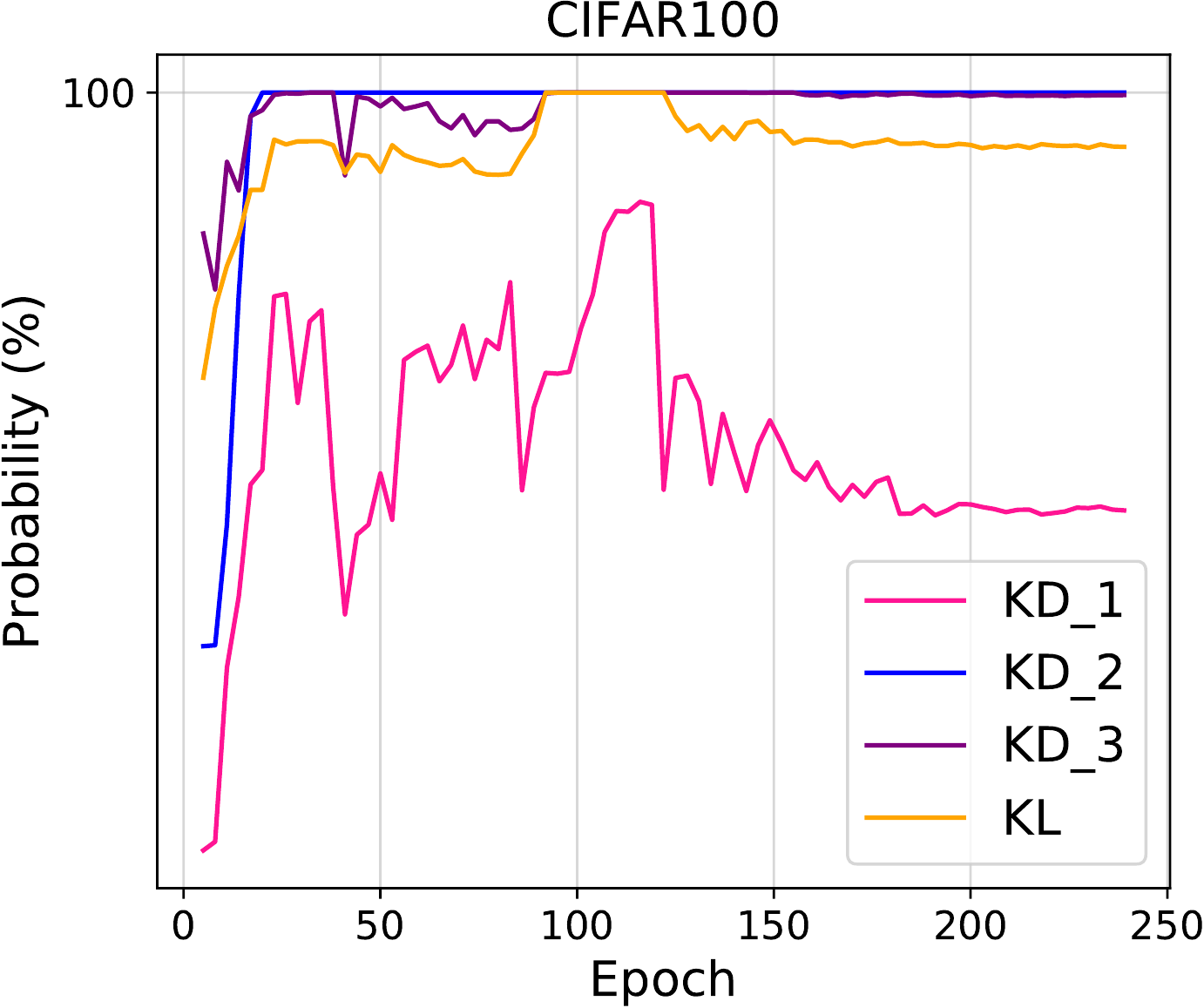}
    \includegraphics[scale=0.293]{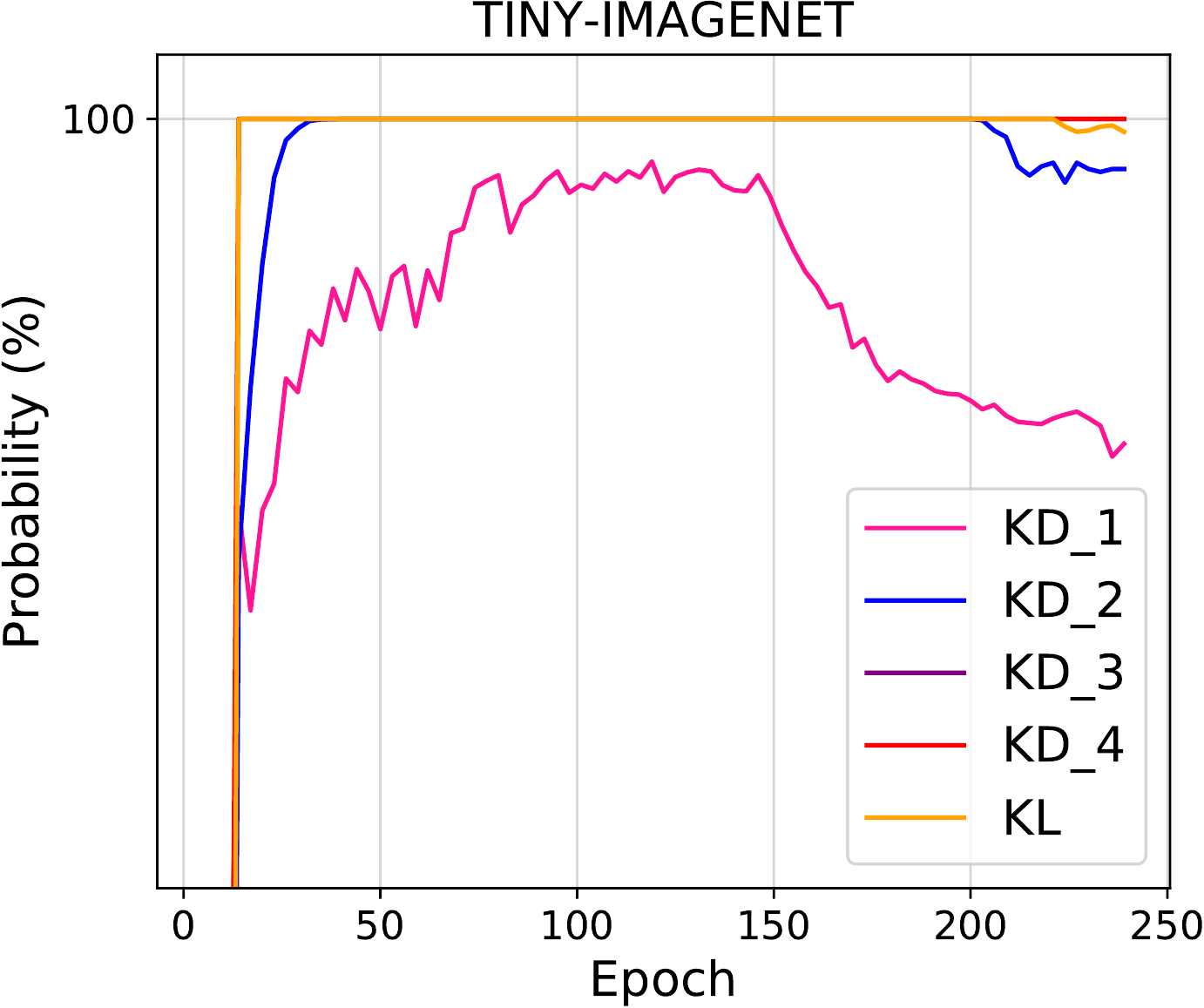}
    \caption{Distillation probability curves at different distillation spots. \textbf{Left}: WRN\_40\_2 $\rightarrow$ WRN\_16\_2, VID~\cite{Ahn2019VariationalID} on CIFAR-100. \textbf{Right}: Vgg13 $\rightarrow$ MobileNetV2, AT~\cite{Zagoruyko2017AT} on tiny-ImageNet. KD\_1, KD\_2, KD\_3, KD\_4, KL are the candidate distillation spots ordered by their position in the network.}
    \label{fig:spots}
\end{figure}

\begin{figure}[ht]
    \centering
    \includegraphics[scale=1.00]{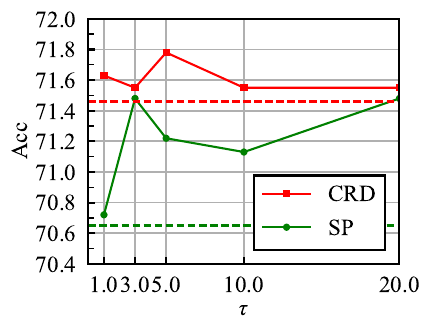}
    \includegraphics[scale=1.00]{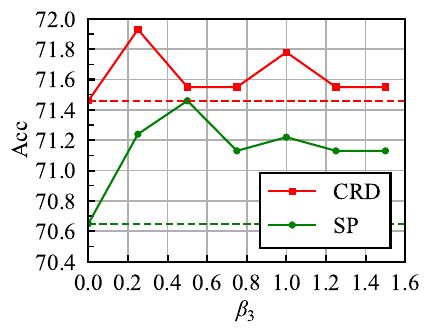}
    \caption{Sensitivity analysis of $\beta_3$ and $\tau$. Experiments are conducted on CIFAR-100. The dotted horizontal lines denote orignial KD without the proposed spot-adaptive distillation.}
    \label{fig:sensitivity}
\end{figure}

\noindent\textit{2) How does the decision change at different spots?}

Here we study how the distillation decisions made by the policy network change at different distillation spots and distillation stages~(\textit{i.e.}, training epochs). At each candidate distillation spot, the probability of distillation is the ratio of the number of samples that are distilled at this spot to the total number of training samples. The probability curves at different spots along the training epochs are depicted in Fig.~\ref{fig:spots}. At early stages, as the teacher network is well trained, the optimal routing decision should be choosing the teacher layers at all branch points in the routing network. Thus the distillation probability should be nearly $100\%$ at all distillation spots. However, as the policy network is randomly initialized and has not been well trained yet, it makes nearly random decisions on routing and thus the distillation probability is low. As the training proceeds, the policy network gradually learns how to make the right decisions and finds that the teacher layers tend to be better, thus the distillation probability increases rapidly. After
a period of distillation, the student model master the knowledge of the teacher. Some samples become less useful for training the student model, and thus the distillation probability decreases~(\textit{e.g.}, KD\_1). Generally speaking, shallow layers are more sensitive to adaptive distillation. Deep layers, on the other hand, requires distillation nearly all the time and for almost all the samples, as shown by curves of KD\_4 and KL. The reasons for this phenomenon may be that the features from shallow layers are relatively noisy for distillation. As the capacity of the student model is much smaller than the teacher, learning from these noisy features degrades its performance on the final target task.
\begin{figure*}[ht]
    \centering
    \includegraphics[scale=0.54]{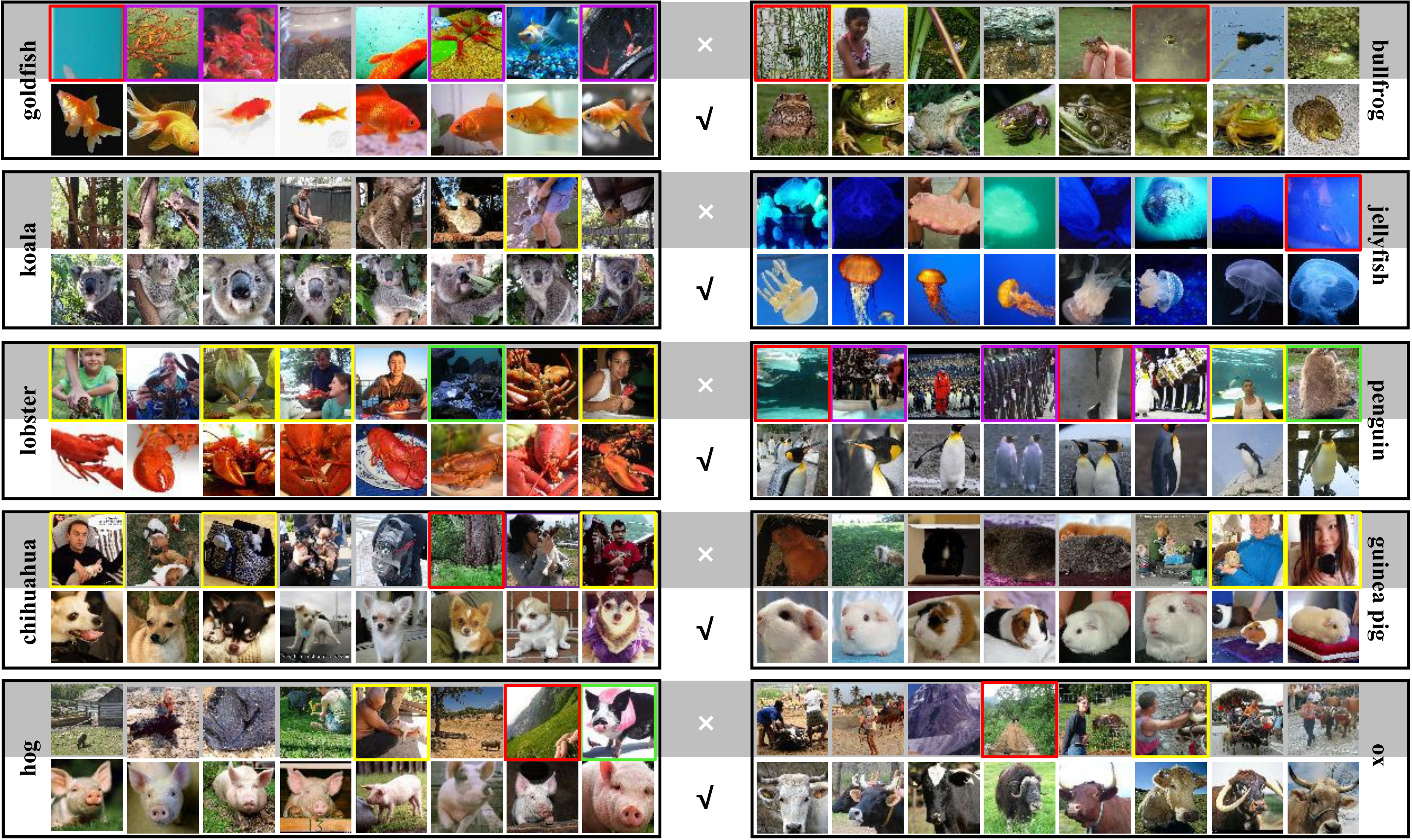}
    \caption{Visualization of the distillation decisions from the proposed SAKD on tiny-ImageNet~\cite{Le2015TinyIV} with VID~\cite{Ahn2019VariationalID} under C-ResNet56 $\rightarrow$ C-ResNet20 distillation. ''$\checkmark$'' denotes distilling knowledge from the teacher to the student, ''$\times$'' means the student will not accept knowledge from the teacher.}
    \label{fig:distillation-vis}
\end{figure*}

\noindent\textit{3) Trainable \textit{vs} frozen teacher networks, which is better?}

The teacher network is frozen all the time in the proposed method. Here we relax this constraint and introduce two alternative settings: (1) the teacher network is randomly initialized and co-trained with the student network; (2) the teacher network is initialized with the pre-trained parameters and co-trained with the student network. The trainable teacher network improves the capacity of the multi-path routing network, but may impair the training of the student model which will be deployed in isolation. Experimental results are provided in Table~\ref{tab:trainable-vs-frozen}. It can be seen that training the teacher network, whether from scratch or from pre-trained parameters, degrades the distillation performance, which verifies our assumption.  What is worse is that training the teacher network slows down the distillation process, as updating teacher parameters needs more computation.
\begin{table}[t]
    \begin{center}
    \resizebox{0.48\textwidth}{!}{
    \begin{tabular}{c|cc|cc}
    \toprule
        \multicolumn{1}{c}{}&\multicolumn{2}{c}{\textbf{PKT}~\cite{Passalis2018LearningDR}} & \multicolumn{2}{c}{\textbf{FitNet}~\cite{Romero2015FitNetsHF}}\\
        \midrule
        \textbf{Teacher} &  C-ResNet56 & Vgg13 & C-ResNet56 & Vgg13\\
        \textbf{Student} & C-ResNet20 &MobileNetV2  &C-ResNet20 & MobileNetV2\\
        \midrule
        Scratch\blue{$^\dag$} & 70.94 & 71.58 & 70.80 & 72.19 \\
        Pre-trained\blue{$^\dag$} & 71.04 & 71.66 & 71.30 & 72.61\\
        SAKD & \textbf{71.56} & \textbf{73.23} & \textbf{71.40} & \textbf{73.11}\\
        \bottomrule
    \end{tabular}
    }
    \end{center}
    \caption{Trainable \textit{vs} frozen teacher network. \blue{$\dag$} denotes the teacher network is co-trained with the student network.}
    \label{tab:trainable-vs-frozen}
\end{table}

\noindent\textit{4) Sensitivity analysis of $\beta_3$ and $\tau$.}

The proposed method involves several hyperparameters. However, most of them are introduced in previous works. We follow the common settings in the literature for these hyperparameters. This work introduces two new hyperparameters, including $\tau$ in Eqn.~\ref{eq:gumbel-softmax} and $\beta_3$ in Eqn.~\ref{eq:overall-loss}. Here we make sensitivity analysis to discuss their impact on the final performance. The results are listed in Fig.~\ref{fig:sensitivity}. It can be seen that both $\beta_3$ and $\tau$ affect the final performance to some degree. However, they make the proposed spot-adaptive method achieve superior results in a wide range of values to standard distillation methods. This feature makes the proposed method very attractive as we do not have make great efforts to tune the parameters.

\noindent\textit{5) Visualization of the distillation decision.}

To better understand the distillation decisions made by the policy network, we provide visualization of distillation decisions on ten categories of tiny-ImageNet~\cite{Le2015TinyIV} in Fig.~\ref{fig:distillation-vis}.
It can be seen that most images that will be distilled ($\checkmark$) are usually of better quality than those not ($\times$).
We summarize the samples without knowledge distillation into four categories: missing content, ambiguous subjects, group of objects and unusual morphology, which are represented by {red}, {yellow}, {purple} and {green} frames in the figure, respectively.

Missing content ({red}). Some images of this type, due to their extreme close-ups or uncharacteristic views, capture only parts of the object, such as the tail of a goldfish, the body of a penguin and the body of a hog. In other images with missing content, the objects are indistinguishable from the background, \textit{e.g.}, bullfrog, jellyfish, penguin, chihuahua and ox.

Ambiguous subjects ({yellow}). These images contain multiple objects without identifying which object is the focus of the image, \textit{e.g.}, bullfrog and human, koala and human leg, lobster and human, penguin and human, chihuahua and human, chihuahua and bag, guinea pig and human, \textit{etc}. With these input images, it is easy for a model to learn features that do not fall into the target category, and eventually leads to errors.

The group of objects ({purple}). The close-up of an individual object reveals its characteristics in detail, whereas the group of samples only provide overall features, \textit{e.g.}, the group of goldfish and the group of penguins.

Unusual morphology ({green}). Some of the images are different from most of those in the dataset, which will not be distilled because of their extraordinary. The rarity of these images makes the features they provide incompatible with the general features. For example, we can see the lobster in blue, the furry penguin and the pink-haired hog in this figure, which provide conflicting features with red lobsters, molted penguins and pigs without head hair that are common in the dataset.

These low-quality features can produce noisy features or predictions, which may hurt the learning of the student model due to its limited capacity. We admit that these non-distilled images can provide information for the model from another perspective, but the noise introduced by them is also worth considering.
Generally, the distillation decision shown in this figure is reasonable as discriminating images offer informative features, and thus the knowledge from these images will guide the student well.

\section{Conclusions}
\label{sec:conclusion}
In this work, we argue knowledge distillation strategy should not be fixed throughout the training process and for all training samples. We thus propose a spot-adaptive distillation strategy to automatically decide on the distillation spots per sample at different distillation stages. The proposed distillation strategy first merges the teacher network and the student network into a multi-path routing network. Then a policy network is introduced to make decisions per sample on the data flow path through the network. The distillation decisions are made based on the routing decisions from the policy network. We evaluate the proposed method under both homogeneous and heterogeneous distillation settings. And extensive ablation study is also conducted to demonstrate the effectiveness of the proposed method. The experimental results showcase that the proposed adaptive distillation strategy consistently improves the performance of various existing distillation methods.

For possible future work, the proposed method can be studied to applied to broader scenarios, \textit{e.g.}, data-free distillation where original training data is not available, or transformer-based distillation where the model architectures are replaced with the trending transformers.

% Can use something like this to put references on a page
% by themselves when using endfloat and the captionsoff option.
\ifCLASSOPTIONcaptionsoff
  \newpage
\fi

\bibliographystyle{IEEEtran}
\bibliography{egbib}
\begin{IEEEbiography}[{\includegraphics[width=1in,height=1.25in,clip,keepaspectratio]{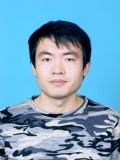}}]{Jie Song}~(Member, IEEE)
is an assistant research fellow in College of Software Technology, Zhejiang University. He received his  B.Sc. degree in Computer Science and Technology from Sichuan University, China, in 2015, and Ph.D. degree in Computer Science and Technology from College of Computer Science, Zhejiang University, China, in 2020. His research interests mainly include knowledge distillation, transfer learning, few- and zero-shot learning, and interpretable machine learning.
\end{IEEEbiography}

\begin{IEEEbiography}[{\includegraphics[width=1in,height=1.25in,clip,keepaspectratio]{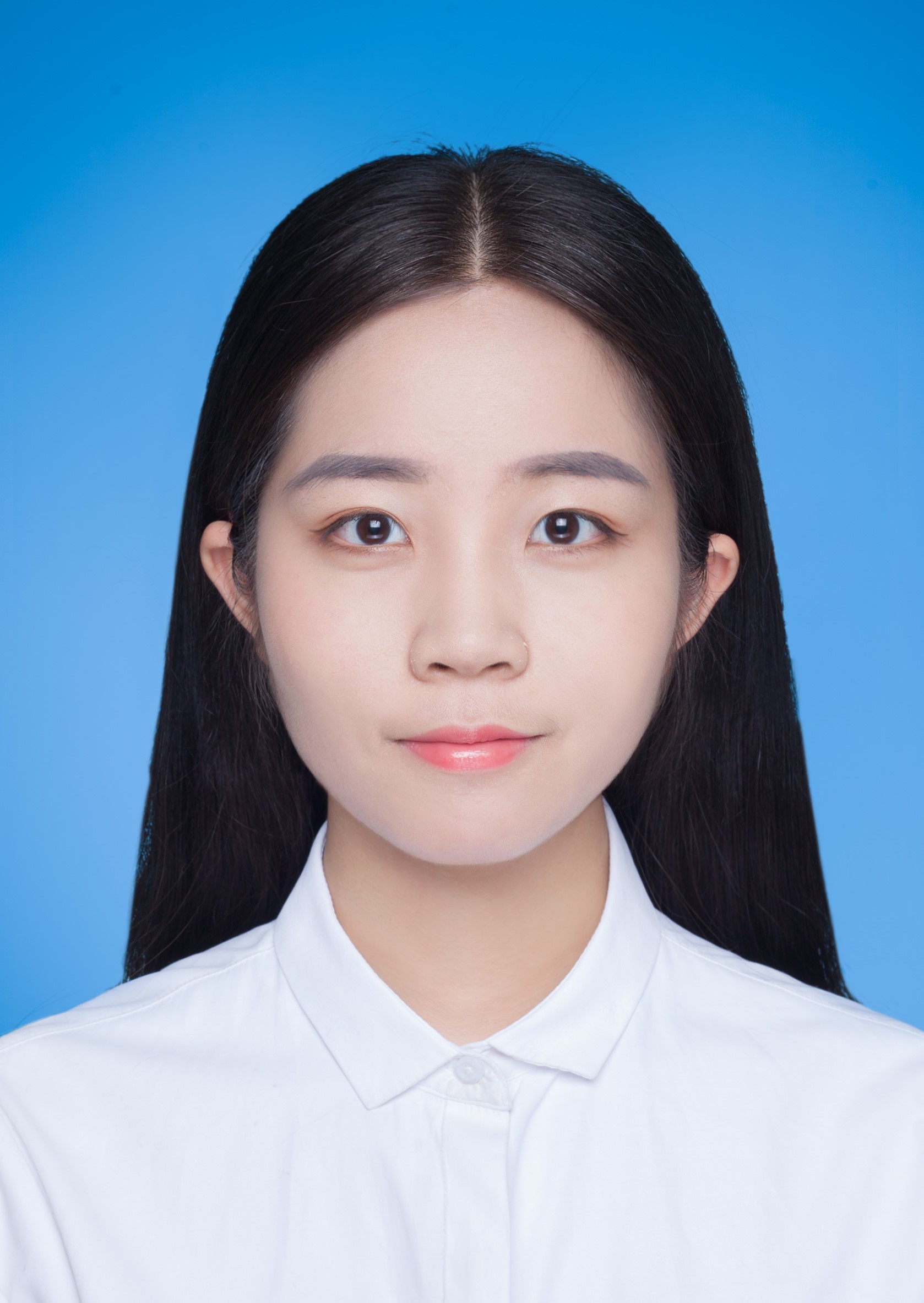}}]{Ying Chen} received the B.E. degree with the College of Software, Zhejiang University. Her research interests include fine-grained image classification and knowledge distillation.
\end{IEEEbiography}

\begin{IEEEbiography}[{\includegraphics[width=1in,height=1.25in,clip,keepaspectratio]{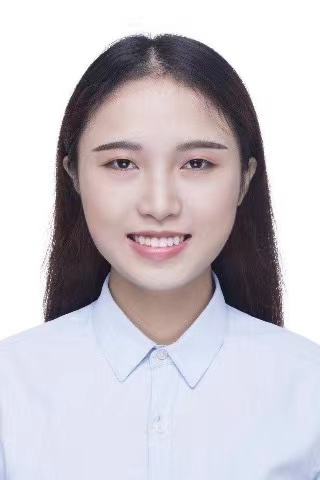}}]{Jingwen Ye}
received the B.Sc. degree from the Dalian University of Technology in 2016 and the Ph.D. degree from Zhejiang University in 2021. She is currently a Postdoctoral Fellow with National University of Singapore. Her research interests include transfer learning, human parsing, scene parsing, depth estimation, and image processing in various applications.
\end{IEEEbiography}

\begin{IEEEbiography}[{\includegraphics[width=1in,height=1.25in,clip,keepaspectratio]{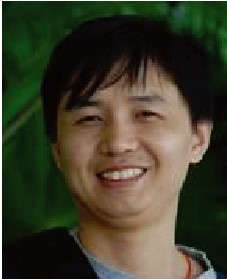}}]{Mingli Song}~(Senior Member, IEEE)
is a Professor in Microsoft Visual Perception Laboratory, Zhejiang University. He received the Ph.D. degree in Computer Science from Zhejiang University, China, in 2006. He was awarded Microsoft Research Fellowship in 2004. His research interests include pattern classification, weakly supervised clustering, color and texture analysis, object recognition, and reconstruction. He is a senior member of the IEEE.
\end{IEEEbiography}

\end{document}